\documentclass{article}
\usepackage{spconf,times}

\usepackage{cite}
\usepackage{amsmath,amssymb,amsfonts}
\usepackage{algorithmic}
\usepackage{graphicx}
\usepackage{textcomp}
\usepackage{xcolor}
\usepackage{booktabs}
\usepackage{url}
\usepackage{multicol,multirow}

\title{DYSHIP: LEVERAGING ACOUSTIC FEATURES FOR SPEAKER
RELATIONSHIP CLASSIFICATION FROM DYADIC CONVERSATIONS}

\name{Kieran Sturges$^{1,2}$, Thomas Thebaud$^{1,3}$,
      Yaohan Guan$^{1}$, Najim Dehak$^{1}$}

\address{
$^{1}$Center for Language and Speech Processing,
Johns Hopkins University, Baltimore, MD, USA\\
$^{2}$CSE and Mathematics Department,
Pennsylvania State University, State College, PA, USA\\
$^{3}$AMIAD, Palaiseau, France
}

\begin{document}
\maketitle

\begin{abstract}
We introduce DyShip, a framework for classifying speaker familiarity and kinship from dyadic conversations in the Seamless Interaction corpus. We compare five speech representations: spectrograms, MFCCs, self-supervised WavLM representations, x-vectors, and estimated articulatory movements. Spectral features perform best for familiarity classification, reaching an accuracy of 87.81\%, while WavLM achieves the highest kinship accuracy of 85.17\%. X-vectors and articulatory features yield weaker results on familiarity classifiaction, but shine for kinship classification, with an AUC of 0.909 for x-vectors. Most classifiers exceed previously reported human and audio language model scores on the same dataset.
These findings highlight the potential of conventional acoustic features for relationship inference and motivate further investigation of the conversational cues captured by different representations.

\end{abstract}


\vspace{-5mm}
\section{Introduction}
\vspace{-3mm}
% why is it an interesting problem?
Spoken dialogue is a rich form of interaction between two individuals; it contains a large set of information beyond its transcript: the identity of speakers, and their individual characteristics, such as their age, gender, and accents, can be predicted from their speech~\cite{feng2025vox, yang2025demographic}. 
Beyond the static characteristics of a speaker, previous works have shown that during oral interactions, one speaker tends to modify their speech to adapt to the other one, both lexically~\cite{brennan1996lexical} and acoustically \cite{gross2013speech}, a phenomenon known as speech entrainment.
Further works have shown that this entrainment is correlated with social relationships between speakers~\cite{bevnuvs2014social}. 
Similarly, individuals of the same family exhibit similar speech patterns: children follow some habits from their parents~\cite{huttenlocher2007varieties}, family groups exhibit coherent syntax~\cite{de2003language}, and even some vocal tract similarities can be measured~\cite{vanderydt1998study}.

% Previous works have shown that the familiarity between two people is a key factor that influences human interaction [2]. For example, strangers use more formal language while familiar dyads use more nonverbal cues [3]. Despite these observable differences, recent work shows that both humans and audio large language models struggle to infer speaker relationships from spoken dyadic conversations [4]. Understanding how speaker relationships are encoded in speech can provide insight into human behavior, with potential applications for socially aware systems. Thus, identifying the properties of speech that explain relationships is an important step toward utilizing relational cues in conversations.

% What is missing?
Building on the recently published \textit{Seamless Interactions} dataset~\cite{agrawal2025seamless}, which contains short spoken dialogues among speakers whose relationship is labeled, recent unpublished works have been looking into predicting the relationship between speakers from the transcript or the spoken interaction~\cite{YAOHAN, girard2026friendbench}
FriendBench~\cite{girard2026friendbench} used the first 20 seconds of dialogues and asked text, audio, and video large language models (LLM) to predict the familiarity of the speakers as a binary output, reaching respectively 56.2, 66.7, and  66.7\% accuracies with their best LLM. Comparatively, individual human raters showed up to 49.9, 56.3, and 60.9\% accuracies, respectively.
Comparatively, \cite{YAOHAN} used the full conversations, up to 5 minutes, and compared different prompting techniques to improve LLM performances on the transcript and the audio, yielding up to 75.0\% accuracy using an audio LLM (Compared to 62.3\% using a zero-shot prompt). With access to the full dialogue, individual annotators obtained a 65.4\% F1 score when given both the full transcript and audio.
As LLM inference is getting increasingly costly and human annotation can be as costly and time-consuming, this pushes forward the interest toward simpler classification systems.

% Existing work on speaker relationship inference uses pretrained speech representations (e.g., WavLM~\cite{chen2022wavlm} and X-vectors~\cite{snyder2018x}) to perform familiar-versus-stranger [5] and kin-versus-nonkin [6] classification. In contrast, traditional spectral representations, such as MFCCs and spectrograms, receive less attention. Furthermore, vocal tract movement has not been investigated, despite its anatomical relevance to kinship [7]. Using abstract representations is an intuitive approach to capturing subtle linguistic, prosodic, and acoustic cues when inferring speaker relationships. However, the relatively moderate performance of existing approaches raises the question of whether such abstraction is necessary, or whether relationship-relevant information is already encoded in acoustic features. Without comparing representations varying in abstraction, it remains unclear what unique information different representations provide for inferring speaker relationships.

% What did you do?
 In this work, we train binary classifiers to predict relationships between 2 speakers given a dialogue.
 We predict both familiar-versus-stranger classification, as defined in the labels of Seamless Interactions~\cite{agrawal2025seamless}, as well as the belonging to ones family, which will be refered as \textbf{kinship}. 
 The classifiers are trained on spectral features (Spectrograms and MFCCs), on Self-Supervised Learning Representations (SSLR, here we use WavLM~\cite{chen2022wavlm}), using vocal tract estimations (predicted by the SPARC system~\cite{cho2024coding}) and x-vectors~\cite{snyder2018x, desplanques2020ecapa}.
 % We compare performances across models and human and large language model benchmarks [9] using two prediction aggregation methods. We further examine prediction-confidence distributions and fusion models to determine whether representations capture overlapping or unique information about speaker relationships. 
% contributions list
Our contributions are threefold:
\vspace{-3mm}
\begin{itemize}
    \item We introduce \textbf{DyShip}, a framework for familiarity and
    kinship classification from dyadic speech, evaluated on
    the Seamless Interaction corpus~\cite{agrawal2025seamless}.
    \vspace{-3mm}
    \item We systematically compare five speech representations,
    showing that conventional spectral features perform best for
    \textit{familiarity} classification, while self-supervised WavLM
    and x-vectors perform best for \textit{kinship} classification.
    \vspace{-3mm}
    \item We provide a reproducible pipeline for feature extraction,
    classifier training, and conversation-level inference, together
    with the code\footnote{\url{https://github.com/kierstur/eDartRelationships}} and the models\footnote{HuggingFace links for the models will be provided upon publication.}.
\end{itemize}

\section{Related Works}

\subsection{Speech Representations}
Speech representations capture complementary aspects of the speech signal, ranging from spectral structure to learned speaker characteristics and articulatory information. 
For a long time, spectral representations in speech processing have been the norm: \textbf{spectrograms} describe the distribution of signal energy across time and frequency, while Mel-frequency cepstral coefficients (\textbf{MFCCs}) provide a compact representation of the spectral envelope, most useful for speaker recognition tasks. 
More recently, self-supervised models such as \textbf{WavLM}~\cite{chen2022wavlm} learn contextual representations from large speech corpora, supporting the extraction of information relevant to multiple speech processing tasks. 
Speaker embeddings, commonly referred to as \textbf{x-vectors}~\cite{snyder2018x}, instead emphasize speaker-discriminative characteristics; in this work, they are extracted using ECAPA-TDNN~\cite{desplanques2020ecapa}. 
Finally, articulatory representations offer a description of speech production: \textbf{SPARC}~\cite{cho2024coding} estimates vocal tract movements from audio, using a fine-tuned WavLM system and a classification head, including the movements of the lips, jaw, and tongue. Comparing these representations allows us to investigate whether speaker relationships are better predicted from conventional acoustic features, general-purpose learned representations, speaker identity cues, or estimated articulatory dynamics.

\subsection{Human and LLM perfomances}
% Our study compares representation-based models to human judgement and zero-shot LLMs. To contextualize our results, we use benchmarks from concurrent work ~\cite{} that evaluates both familiar-versus-stranger and kin-versus-nonkin classification using conversational audio from the Seamless Interaction dataset. By covering both speaker relationship inference tasks with the same data source, these results provide consistent benchmarks for interpreting the performance of our representation-based models.

Recent studies have investigated how accurately humans and large language models can infer speaker relationships from conversational cues. FriendBench~\cite{girard2026friendbench} evaluates 26 models from seven companies on familiarity classification using 20-second excerpts from 96 balanced dyads in the Seamless Interaction dataset. Its comparison of text, audio, and video inputs highlights the contribution of non-textual cues, with audio-based predictions outperforming those based on transcripts alone. The study also reports a tendency for models to predict strangers, whereas human judgments are more balanced. Complementing this short-excerpt setting, concurrent work~\cite{YAOHAN} examines relationship inference using full conversations and different prompting strategies, providing additional human and LLM reference results. These studies motivate our investigation of acoustic representations by showing that conversational speech contains cues relevant to relationship inference, while leaving open which representations capture them most effectively. We use their reported results to contextualize our classifiers, while recognizing differences in input duration, supervision, and evaluation protocols.

\section{Materials and Methods}

\subsection{Dataset}
We use the Seamless Interaction dataset~\cite{agrawal2025seamless}, which contains over 4,000 hours of recorded dyadic interactions from over 4,000 participants. The dataset includes diversity in participant age, race, gender, and education level, and provides a range of naturalistic and acted speech interactions, guided by conversational prompts. The naturalistic subset also contains interpersonal relationships labels for 11 relation classes: \textit{'strangers'} are \textbf{neither familiar nor kinship}, \textit{'friends'}, \textit{'coworkers'}, \textit{'roommates'}, \textit{'romantic partners'}, \textit{'classmates'}, \textit{'neighbors'}, and \textit{'generic familiar'} are defined as \textbf{familiar relations but non-kin}, while \textit{'siblings'}, \textit{'parent-child'} pairs, and \textit{'generic family'} are both \textbf{kinship and familiar}.
We used the train, development, and test splits provided by Seamless Interaction to ensure that no participant is shared between splits. 
For the training and development sets, we only kept utterances that were at least five seconds long, and segmented longer utterances into five second segments. 
In contrast, the test set included all utterances to match previous evalutation frameworks. 
We also excluded interactions that did not involve exactly two participants or for which the interpersonal relationship label was not in the metadata. Lastly, interactions missing dyad IDs from the metadata were retained only in the training set.
Table \ref{tab:dataset} summarizes the sizes of the resulting datasets after applying these criteria.
The final dataset contains 5.14\% conversations from kin dyads, and 51.00\% from familiar dyads.
% \begin{table}[h]
% \caption{Duration and unique dyads per subset, as well as utterances and conversation counts, and the proportion of familiar and kinship samples.}
% \label{tab:dataset}
% \centering
% \resizebox{\linewidth}{!}{%
% \begin{tabular}{lll lll lll}
% \toprule
% \multirow{2}{*}{\textbf{Dataset}} & \multirow{2}{*}{\textbf{Dur.}} & \textbf{Num.} & \multicolumn{3}{c}{\textbf{Utterances}} & \multicolumn{3}{c}{\textbf{Conversations}} \\
% & & \textbf{Dyads} & Number & \% Fam. & \% Kin & Number & \% Fam. & \% Kin \\
% \midrule
% Train & 579.59h & 3118 & 417,306 & 50.45\% & 4.57\% & 41,852 & 21,135 & 2,052 \\
% Dev   & 6.36h & 47 & 4,577 & 60.84\% & 11.0\% & 559 & 356 & 60 \\
% Test  & 27.10h & 55 & 17,214 &  72.76\% & 13.57\% & 625 & 464 & 99 \\
% \midrule
% Total & \\
% \bottomrule
% \end{tabular}
% }
% \end{table}

\begin{table}[h]
\caption{Durations and count of speakers, unique dyad, conversations, and utterances per subset.}
\label{tab:dataset}
\centering
\resizebox{\linewidth}{!}{%
\begin{tabular}{lll ll ll}
\toprule
\textbf{Set} & \textbf{Duration} & \textbf{Speakers} & \textbf{Dyads} & \textbf{Conv.} & \textbf{Utt.} \\
& (h) & $S_i$ & $D_{i,j}$ & $C_{i,j}$ & $U_{C_{i,j}}^n$ \\
\midrule
Train & 579.58 & 431 & 3,118 & 41,852 & 417,306 \\
Dev   &   6.36 &   103 &    47 &    559 &   4,577 \\
Test  &  28.03 &   103 &    55 &    607 &  28,220 \\
\midrule
Total & 613.97 & 637 & 3,220 & 43,018 & 450,103 \\
\bottomrule
\end{tabular}
}
\end{table}

\subsection{Models}
% What's the general architecture?
% Features -> Classifier -> output

%The pipeline then encodes the audio into a feature representation (either frozen or unfrozen) and passes this representation through a classifier to generate a prediction. For each feature representation, we developed two separate models: one for familiar-versus-stranger binary classification and one for kin-versus-nonkin binary classification.

\begin{figure*}[ht]
    \centering
    \vspace{-6mm}
    \includegraphics[width=0.9\linewidth]{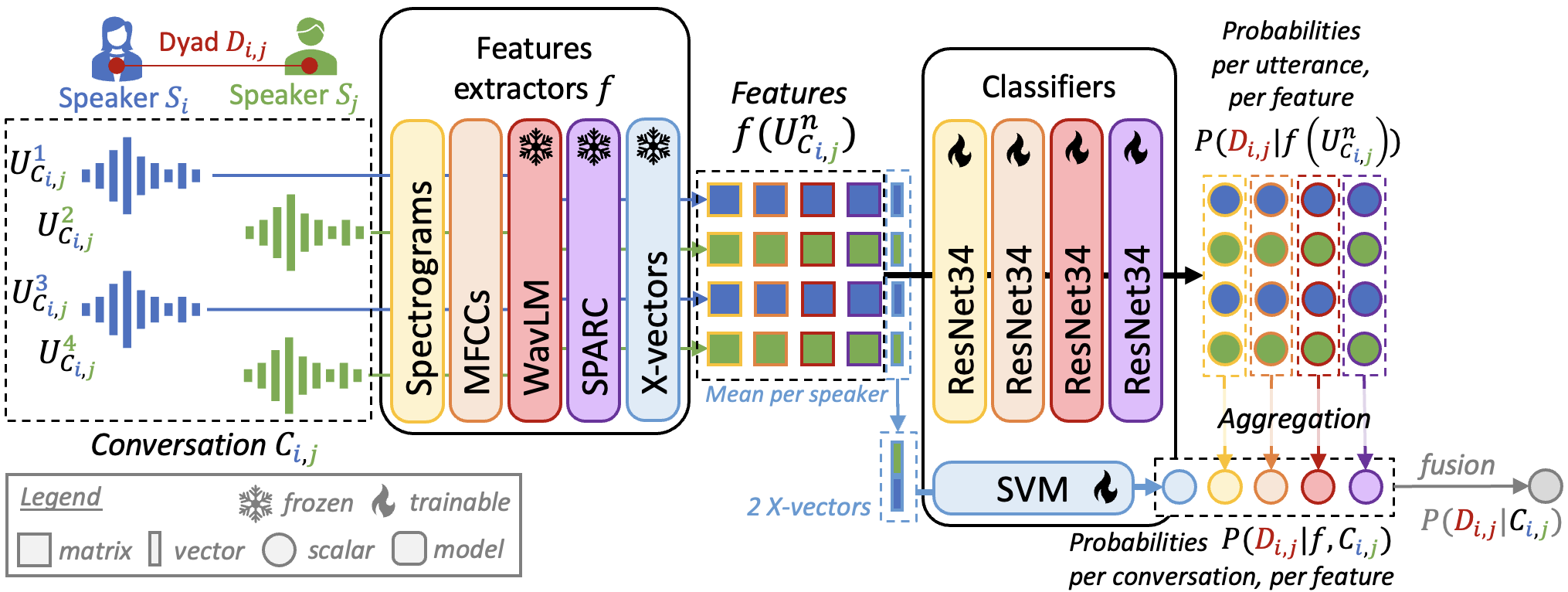}
    \caption{Schematic of the pipeline: Each utterance is processed separately by different feature extractor, then by a dedicated classifier to predict the probability per utterance, aggregated into a probability per conversation. }
    \label{fig:model}
    \vspace{-4mm}
\end{figure*}

\subsubsection{Features}
\label{ssec:features}
We train one binary classifier per feature and per task, considering five representations: 
%two spectral representations and three pretrained neural representations. %These representations provide a range of acoustic and higher-level features for comparison.
\textbf{Spectral}: MFCCs and spectrograms are extracted using Pytorch~\cite{imambi2021pytorch}. 
\textbf{SSLR}: We extract representations from the final layer of the frozen WavLM-base-plus~\cite{chen2022wavlm} model, yielding 768 dimension embeddings.
\textbf{Vocal Tract}: The pretrained SPARC model~\cite{cho2024coding} is used to predict 6 vocal tract movements in the sagittal plane across six articulators: the upper lip, lower lip, lower incisor, tongue tip, tongue blade, and tongue dorusm.
\textbf{x-vectors} are extracted using the pretrained ECAPA-TDNN~\cite{desplanques2020ecapa} from Speechbrain~\cite{speechbrain}.

% \paragraph{Spectral features}
% We use MFCCs [22] and spectrograms [23] as our spectral features. 
% MFCCs provide compact, short-term phonetic and vocal tract information, while spectrograms retain more detailed frequency information over time. 

% \paragraph{WavLM}
% % to cite something:
% We use a frozen WavLM-base-plus~\cite{chen2022wavlm} model. We extract representations from the final layer of the model to capture many higher-level properties of speech used in many speech-related tasks.

% \paragraph{SPARC}
% We use frozen SPARC [24] vectors to represent vocal tract movements in the sagittal plane across six articulators: the upper lip, lower lip, lower incisor, tongue tip, tongue blade, and tongue dorusm. These representations provide a unique articulatory characterization of speech that differs from standard acoustic approaches.

% \paragraph{X-Vector systems}
% We use frozen X-Vector [25] embeddings, specifically the ECAPA-TDNN architecture, to obtain learned representations commonly used for speaker verification. These embeddings allow us to evaluate whether speaker-related information contributes to speaker relationship inference tasks.

\subsubsection{Classifiers}
\label{ssec:classifiers}

For MFCC, spectrogram, WavLM, and SPARC representations, we used a ResNet-34 classifier~\cite{he2016deep} with a binary classifier head.
For X-Vector representations, we used a non-linear support vector machine (SVM) classifier from scikit-learn package ~\cite{kramer2016scikit}. The X-Vector representations are standardized using StandardScaler before classification, and we use a radial basis function (RBF) kernel, set the C value to 1, set the gamma value to scale, and balanced class weights. We also used this classifier architecture for fusion models.

\vspace{-3mm}
\subsection{Metrics}
We evaluated each model at the conversation level using unweighted accuracy, macro f1 score, and the area under the receiver operating characteristic curve (AUC). 
All the metrics are evaluated on the conversation-level probabilities $P(D_{i,j}|f, C_{i,j})$.

\vspace{-3mm}
\subsection{Experimental setup}
A dyad $D_{i,j}$ is defined as a set of 2 speakers $S_i$ and $S_j$ ($i\ne j$), which has a fixed \textit{familiarity} (familiar vs stranger) and a fixed \textit{kinship} (kin vs non-kin).
Given each conversation $C_{i,j}$ is defined as a sequence of $N$ utterances $U_{C_{i,j}}^n, n \in [\![1,N]\!]$ from 2 speakers $S_i$ and $S_j$.

\vspace{-3mm}
\subsubsection{Feature extractors}
First, each utterance $U_{C_{i,j}}^n$ is processed to extract a set of features $f(U_{C_{i,j}}^n)$ from a frozen feature extractor $f$, as described in section \ref{ssec:features}.

\vspace{-3mm}
\subsubsection{ResNet classifiers training}
One classifier is trained per type of feature extracted and per task (familiar-vs-stranger and kin-vs-nonkin).
The ResNet34 classifiers are trained with a batch size of 32, learning rate of 0.0001, class-weighted cross-entropy loss, and the Adam optimizer. 
The classifiers are trained for 10 epochs on a single Nvidia 1080Ti GPU (X hours of training time); the checkpoint with the best validation loss is selected for evaluation.
The familiarity probabilities are extracted for each utterance in the test set to obtain the familiarity probability $P$ of a dyad $D_{i,j}$ given the observation of one utterance $P(D_{i,j}|f(U_{C_{i,j}}^n))$.

\subsubsection{Conversation-level aggregation}
Once the utterance-wise predictions $P(D_{i,j}|f(U_{C_{i,j}}^n))$ have been computed for the full test set, they are aggregated using a weighted mean to obtain one probability per conversation $C_{i,j}$:
\vspace{-3mm}
\begin{equation}
    P(D_{i,j}|f, C_{i,j}) = \frac{\sum_{n=1}^{N}\alpha_n{P(D_{i,j}|f(U_{C_{i,j}}^n))}}{\sum_{n=1}^{N}\alpha_n}
\end{equation}
We explore 2 options for the weights $\alpha_n$: Unweighted fusion ($\forall n, \alpha_n=1$) and length-condition fusion, so longer utterances have a larger weight in the sum ($\forall n, \alpha_n=len(U_{C_{i,j}}^n)$).

\subsubsection{The X-vector exception}
The x-vectors are processed differently than the rest of the features:
one x-vector is extracted per utterance, then the x-vectors belonging to one speaker in a conversation are pooled using a mean to get one x-vector per speaker. Interactions missing a dyad ID assume that all utterances belong to one speaker, and the pooled x-vector is duplicated.
Finally, an SVM model is trained to predict the familiarity of the conversation directly from the tuple of x-vectors.
SVMs are trained on CPUs using the scikit-learn package~\cite{kramer2016scikit}.

% \subsubsection{Cross-feature correlation}
\subsubsection{Features fusions}
Finally, a fusion of all features is learned as a weighted sum of the conversation-level probabilities $P(D_{i,j}|f, C_{i,j})$ for all features $f \in \{$Spectrogram, MFCC, SPARC, WavLM, X-vector$\}$.
% \begin{equation}
%     P(D_{i,j}|C_{i,j}) = \frac{\sum_{f}{\beta_f P(D_{i,j}|f, C_{i,j})}}{\sum_{f}{\beta_f}}
% \end{equation}
The weights are learned either using a linear SVM to maximize the accuracy on the dev set, or all fixed at 1 for unweighted fusion.
% \begin{table}[h]
% \caption{Ground Truth Conversation Distribution}
% \label{tab:dataset}
% \centering
% \begin{tabular}{lrrrrr}
% \hline
% \textbf{Dataset}  & \textbf{Conversations} & \textbf{Dyads} & \textbf{Familiar (Convs)} & \textbf{Kin (Convs)} \\
% \hline
% Train & 41,852 & 3118 & 21,135 & 2,052 \\
% Dev   & 559 & 47 & 356 & 60 \\
% Test  & 625 & 55 & 464 & 99 \\
% \hline
% \end{tabular}
% \end{table}

\section{Results}

Table \ref{tab:results} shows the classification results shown by each classifier as well as their fusion, compared with existing studies.

\textbf{Familiar-vs-Stranger Classification:} Most classifiers show a significant improvement both from human capabilities and audio LLMs.
The ResNet34 classifiers showed up to 86.49\% of accuracy for the Spectrograms and up to 0.917 AUC for the MFCCs, leading the way toward low-ressource high-perfoming systems.
The SPARC features showed second-to-last performances, probably explained by the speaker-wise normalization of the values by the model, which would have erased some of the intrinsic properties of the speech related to each speaker.
However, they seems to gain the most from the aggregation by length of the utterances.
The X-vector-based system performed the worst.

\textbf{Kin-vs-Nonkin Classification:} The kinship classification exhibited a much lower range of performances, led by the WavLM-based system with an 85.17\% accuracy.
X-vectors-based and SPARC-based systems showed a higher discrimination on that task, with up to 0.909 AUC for the x-vectors.
This may be explained 

% \textbf{Independence of the predictions:} To motivate the idea of a fusion between the different fusion systems, we measure the pearson correlation between the per-conversation probabilities extracted from each features, and report their values in Figure \ref{fig:correlations}.
% All pairs of features have been found to be statistically significantly correlated to each other (p-values$\leq0.05$), but to varying degrees.
% As expected, spectral features (Spectrograms compared to MFCCs) show the higher correlation at 0.87, joined closely by WavLM features. The SPARC features, despite being predicted from WavLM outputs, show the least correlation with any other set of features.
% The relatively low correlation between the SPARC features, the x-vectors and the other ones shows that each set of features learns different aspects of the relationships to classify.
% This motivates our idea to learn a fusion from all the classifiers.

\textbf{Cross-feature fusion:} The weighted fusion brings almost no improvement, which could be explained by the domain difference between the dev and the test sets: the dev set evaluates on 5-second utterances while the test set uses all of them.
Contrastively, the unweighted fusion shows the best performances across almost all metrics, with up to 0.925 AUC for familiar classification and 0.916 AUC for kinship.

\begin{table}[t]
\centering
\caption{Accuracy (Acc), Macro F1 scores (F1), and AUC metrics across representations and aggregation methods at the conversation level.}
\label{tab:results}
\resizebox{\linewidth}{!}{%
\begin{tabular}{ll|ccc|ccc}
\toprule
\textbf{Feature} & \textbf{System} & \multicolumn{3}{c|}{\textbf{Familiar-vs-Stranger}} & \multicolumn{3}{c}{\textbf{Kin-vs-Nonkin}} \\
& & Acc. & F1 & AUC & Acc. & F1 & AUC \\
\midrule
\multicolumn{8}{c}{Unweighted Average of the utterances} \\
\midrule
MFCC & ResNet34         & 67.38 & 66.33 & \textbf{0.917} & 84.02 & 68.51 & 0.849 \\
Spectrogram & ResNet34  & \textbf{86.49} & 82.09 & 0.897 & 82.87 & 72.11 & 0.875 \\
WavLM & ResNet34        & 85.67 & \textbf{82.20} & 0.910 & \textbf{85.01} & \textbf{73.70} & 0.891 \\
SPARC & ResNet34        & 65.73 & 64.59 & 0.828 & 84.84 & 61.41 & 0.805 \\
X-Vector & SVM          & 65.57 & 60.58 & 0.684 & 83.36 & 45.46 & \textbf{0.909} \\

\midrule
\multicolumn{8}{c}{Weighted Average by length of the utterances} \\
\midrule
MFCC & ResNet34         & 76.61 & 74.59 & \textbf{0.917} & 81.71 & 69.30 & 0.838 \\
Spectrogram & ResNet34  & \textbf{87.81} & \textbf{83.76} & 0.899 & 81.22 & 70.76 & 0.873 \\
WavLM & ResNet34        & 84.51 & 81.89 & \textbf{0.917} & \textbf{85.17} & \textbf{73.69} & 0.885 \\
SPARC & ResNet34        & 71.33 & 69.50 & 0.846 & 85.01 & 69.33 & 0.818 \\
X-Vector & SVM          & 65.57 & 60.58 & 0.684 & 83.36 & 45.46 & \textbf{0.909} \\
\midrule
\multicolumn{2}{l|}{Weighted Fusion by SVM} & 80.41 & 76.75 & 0.893 & 76.94 & 78.07 & 0.798 \\
\multicolumn{2}{l|}{Unweighed Fusion} & 86.01 & 86.87 & 0.925 & 88.96 & 87.27 & 0.916 \\

\midrule
\multicolumn{8}{c}{From the whole conversation} \\
\midrule
% X-Vector & SVM & 62.85 & 59.41 & 0.670 & 49.81 & 45.62 & 0.606 \\
Audio & Best LLM~\cite{YAOHAN}      & 74.0 & 50.8 & - & \\
Audio+Transcript & Best LLM~\cite{YAOHAN} & 75.0 & 66.5 & - & \\
Audio & Human~\cite{YAOHAN}        & 67.8 & 62.8 & - & \\
Audio+Transcript & Human~\cite{YAOHAN}   & 72.2 & 65.4 & - & \\
\midrule
\multicolumn{8}{c}{From the first 20 seconds} \\
\midrule
Audio & Best LLM~\cite{girard2026friendbench}       & 66.7 & - & - & \\
% Audio & Human(indiv.)~\cite{girard2026friendbench} & 56.3 & - & - & \\
Audio & Human~\cite{girard2026friendbench}          & 63.5 & - & - & \\
\bottomrule
\end{tabular}%
}
\vspace{-4mm}
\end{table}

% \begin{figure}[ht]
%     \centering
%     \includegraphics[width=\columnwidth]{Figures/Histogram-Probability.png}
%     \caption{Histograms of models evaluated under the probability aggregation method. The top row shows familiarity, the bottom row shows kinship.}
%     \label{fig:histograms}
% \end{figure}

% \begin{figure}[ht]
%     \centering
%     \includegraphics[width=\columnwidth]{Figures/Probability-Pearson-Heatmap.png}
%     \caption{Heatmap of Pearson correlation coefficients measured between conversation-wise probabilities.}
%     \label{fig:correlations}
% \end{figure}

\vspace{-4mm}
\section{Discussion and Conclusion}
This article presents DyShip, a study of whether acoustic information from dyadic conversations can reveal speaker familiarity and kinship. Using the Seamless Interaction dataset, we compare five speech representations: MFCCs, spectrograms, WavLM embeddings, estimated vocal tract movements, and x-vectors. Among individual representations, spectral features provide the strongest familiarity results, with MFCCs achieving an AUC of 0.917 and spectrograms reaching an F1 score of 83.76\%. For kinship, WavLM achieves the highest accuracy (85.17\%) and F1 score (73.69\%), while x-vectors yield the highest AUC (0.909). Unweighted fusion further improves F1 and AUC for both tasks, reaching 86.87\% and 0.925 for familiarity, and 87.27\% and 0.916 for kinship, outperforming SVM-based fusion on all reported metrics. These results support acoustic representations as effective baselines for relationship inference, although comparisons with reported human and audio LLM scores remain subject to differences in evaluation protocols.

Despite these results, several limitations remain. Classification is \textit{evaluated on a single dataset}, with only 47 and 55 unique dyads in the development and test sets, respectively. Although participants are disjoint across splits, multiple conversations per dyad produce dependent observations and limit generalization. Kinship classification also exhibits \textit{substantial class imbalance}, making accuracy alone insufficient to characterize performance. The x-vector system illustrates this discrepancy, achieving an AUC of 0.909 but an F1 score of only 45.46\% at the evaluated decision threshold. Moreover, utterance-level classification followed by aggregation provides \textit{no explicit modeling of turn-taking, cross-speaker adaptation, or temporal dependencies}. Finally, correlations between predictions do not identify the acoustic cues captured by each representation. Although unweighted fusion improves performance, the weaker SVM-based fusion results show that combining representations does not necessarily yield gains.

Future work will investigate explicit modeling of cross-speaker dynamics by processing complete conversations and extend evaluation to other datasets with relationship annotations, such as CallHome~\cite{canavan1997callhome, katerenchuk2018interpersonal}. Additional data could broaden the diversity of evaluated relationships and support a more balanced assessment of kinship classification.

% \newpage
% \section{Aknowlegments}
% This work has been made possible thanks to the PROPEL program, edition 2026.

\bibliographystyle{IEEEbib}
\bibliography{main}

@article{bevnuvs2014social,
  title={Social aspects of entrainment in spoken interaction},
  author={Be{\v{n}}u{\v{s}}, {\v{S}}tefan},
  journal={Cognitive Computation},
  volume={6},
  number={4},
  pages={802--813},
  year={2014},
  publisher={Springer}
}

@article{brennan1996lexical,
  title={Lexical entrainment in spontaneous dialog},
  author={Brennan, Susan E and others},
  journal={Proceedings of ISSD},
  volume={96},
  pages={41--44},
  year={1996}
}

@article{canavan1997callhome,
  title={Callhome american english speech},
  author={Canavan, Alexandra and Graff, David and Zipperlen, George},
  journal={Linguistic Data Consortium},
  year={1997}
}

@misc{YAOHAN,
      title={Who Are They to Each Other? Multi-Agent Reasoning for Speaker Relationship Inference}, 
      author={Yaohan Guan and Yen-Ju Lu and Yuzhe Wang and Junhyeok Lee and Jesus Villalba and Laureano Moro Velazquez and Thomas Thebaud and Najim Dehak},
      year={2026},
      eprint={2609.09628},
      archivePrefix={arXiv},
      primaryClass={cs.MA},
      url={https://arxiv.org/abs/2609.09628}, 
}

@inproceedings{katerenchuk2018interpersonal,
  title={Interpersonal Relationship Labels for the CALLHOME Corpus},
  author={Katerenchuk, Denys and Brizan, David Guy and Rosenberg, Andrew},
  booktitle={Proceedings of the Eleventh International Conference on Language Resources and Evaluation (LREC 2018)},
  year={2018}
}

@article{cho2024coding,
  title={Coding speech through vocal tract kinematics},
  author={Cho, Cheol Jun and Wu, Peter and Prabhune, Tejas S and Agarwal, Dhruv and Anumanchipalli, Gopala K},
  journal={IEEE Journal of Selected Topics in Signal Processing},
  volume={18},
  number={8},
  pages={1427--1440},
  year={2024},
  publisher={IEEE}
}

@incollection{kramer2016scikit,
  title={Scikit-learn},
  author={Kramer, Oliver},
  booktitle={Machine learning for evolution strategies},
  pages={45--53},
  year={2016},
  publisher={Springer}
}

@inproceedings{he2016deep,
  title={Deep residual learning for image recognition},
  author={He, Kaiming and Zhang, Xiangyu and Ren, Shaoqing and Sun, Jian},
  booktitle={Proceedings of the IEEE conference on computer vision and pattern recognition},
  pages={770--778},
  year={2016}
}

@incollection{imambi2021pytorch,
  title={PyTorch},
  author={Imambi, Sagar and Prakash, Kolla Bhanu and Kanagachidambaresan, GR},
  booktitle={Programming with TensorFlow: solution for edge computing applications},
  pages={87--104},
  year={2021},
  publisher={Springer}
}

@article{de2003language,
  title={Language variation and local elements in family discourse},
  author={De Houwer, Annick},
  journal={Language variation and change},
  volume={15},
  number={3},
  pages={329--349},
  year={2003},
  publisher={Cambridge University Press}
}

@article{girard2026friendbench,
  title={FriendBench: Benchmarking Dyadic Familiarity Inference in Humans and Multimodal Large Language Models},
  author={Girard, Jeffrey M and Zheng, Jason Z and Vertino, Jacqueline R and D'Avirro, Antony and Peloquin, Benjamin},
  journal={arXiv preprint arXiv:2607.29602},
  year={2026}
}

@article{agrawal2025seamless,
  title={Seamless interaction: Dyadic audiovisual motion modeling and large-scale dataset},
  author={Agrawal, Vasu and Akinyemi, Akinniyi and Alvero, Kathryn and Behrooz, Morteza and Buffalini, Julia and Carlucci, Fabio Maria and Chen, Joy and Chen, Junming and Chen, Zhang and Cheng, Shiyang and others},
  journal={arXiv preprint arXiv:2506.22554},
  year={2025}
}

@book{vanderydt1998study,
  title={A study of voice similarities among family members},
  author={Vanderydt, Rachel Evan},
  year={1998},
  publisher={MGH Institute of Health Professions}
}

@article{gross2013speech,
  title={Speech rhythms and multiplexed oscillatory sensory coding in the human brain},
  author={Gross, Joachim and Hoogenboom, Nienke and Thut, Gregor and Schyns, Philippe and Panzeri, Stefano and Belin, Pascal and Garrod, Simon},
  journal={PLoS biology},
  volume={11},
  number={12},
  pages={e1001752},
  year={2013},
  publisher={Public Library of Science San Francisco, USA}
}

@article{huttenlocher2007varieties,
  title={The varieties of speech to young children.},
  author={Huttenlocher, Janellen and Vasilyeva, Marina and Waterfall, Heidi R and Vevea, Jack L and Hedges, Larry V},
  journal={Developmental psychology},
  volume={43},
  number={5},
  pages={1062},
  year={2007},
  publisher={American Psychological Association}
}

@article{chen2022wavlm,
  title={Wavlm: Large-scale self-supervised pre-training for full stack speech processing},
  author={Chen, Sanyuan and Wang, Chengyi and Chen, Zhengyang and Wu, Yu and Liu, Shujie and Chen, Zhuo and Li, Jinyu and Kanda, Naoyuki and Yoshioka, Takuya and Xiao, Xiong and others},
  journal={IEEE Journal of Selected Topics in Signal Processing},
  volume={16},
  number={6},
  pages={1505--1518},
  year={2022},
  publisher={IEEE}
}

@inproceedings{snyder2018x,
  title={X-vectors: Robust dnn embeddings for speaker recognition},
  author={Snyder, David and Garcia-Romero, Daniel and Sell, Gregory and Povey, Daniel and Khudanpur, Sanjeev},
  booktitle={2018 IEEE international conference on acoustics, speech and signal processing (ICASSP)},
  pages={5329--5333},
  year={2018},
  organization={IEEE},
doi = {10.1109/ICASSP.2018.8461375}
}

@misc{speechbrain,
  title={{SpeechBrain}: A General-Purpose Speech Toolkit},
  author={Mirco Ravanelli and Titouan Parcollet and Peter Plantinga and Aku Rouhe and Samuele Cornell and Loren Lugosch and Cem Subakan and Nauman Dawalatabad and Abdelwahab Heba and Jianyuan Zhong and Ju-Chieh Chou and Sung-Lin Yeh and Szu-Wei Fu and Chien-Feng Liao and Elena Rastorgueva and François Grondin and William Aris and Hwidong Na and Yan Gao and Renato De Mori and Yoshua Bengio},
  year={2021},
  eprint={2106.04624},
  archivePrefix={arXiv},
  primaryClass={eess.AS},
  note={arXiv:2106.04624}
}

@article{desplanques2020ecapa,
  title={Ecapa-tdnn: Emphasized channel attention, propagation and aggregation in tdnn based speaker verification},
  author={Desplanques, Brecht and Thienpondt, Jenthe and Demuynck, Kris},
  journal={arXiv preprint arXiv:2005.07143},
  year={2020}
}

@inproceedings{yang2025demographic,
  title={Demographic Attributes Prediction from Speech Using WavLM Embeddings},
  author={Yang, Yuchen and Thebaud, Thomas and Dehak, Najim},
  booktitle={2025 59th Annual Conference on Information Sciences and Systems (CISS)},
  pages={1--6},
  year={2025},
  organization={IEEE}
}

@article{feng2025vox,
  title={Vox-Profile: A Speech Foundation Model Benchmark for Characterizing Diverse Speaker and Speech Traits},
  author={Feng, Tiantian and Lee, Jihwan and Xu, Anfeng and Lee, Yoonjeong and Lertpetchpun, Thanathai and Shi, Xuan and Wang, Helin and Thebaud, Thomas and Moro-Velazquez, Laureano and Byrd, Dani and others},
  journal={arXiv preprint arXiv:2505.14648},
  year={2025}
}

\end{document}